\documentclass{elsarticle}
\makeatletter
\def\ps@pprintTitle{%
	\let\@oddhead\@empty
	\let\@evenhead\@empty
	\def\@oddfoot{}%
	\let\@evenfoot\@oddfoot}
\makeatother

\usepackage{amsmath,graphicx}
\usepackage{multirow}
\usepackage{times}
\usepackage{amssymb}
\usepackage[utf8]{inputenc}
\usepackage{url}
\usepackage{subfig}
\usepackage{array}
\usepackage{float}
\usepackage{grffile}
\usepackage{rotating}
\usepackage{color}
\usepackage{afterpage}
\usepackage{wrapfig}
\usepackage{makecell}
\usepackage{siunitx}
\usepackage[pdf]{pstricks}
\usepackage{makecell}
\usepackage{pgfplots}

\biboptions{authoryear}

\begin{document}
	
	\begin{frontmatter}
		
		\author[lfb,cor]{Michael Gadermayr}
		\author[lfb]{Ann-Kathrin Dombrowski}
		\author[neuro]{Barbara Mara Klinkhammer}
		\author[neuro]{Peter Boor}
		\author[lfb]{Dorit Merhof}
		\address[lfb]{Institute of Imaging \& Computer Vision, RWTH Aachen University, Aachen, Germany}
		\address[neuro]{Institute of Pathology, University Hospital Aachen, RWTH Aachen University, Aachen, Germany}
		\address[cor]{Corresponding author: M. Gadermayr, \url{Michael.Gadermayr@lfb.rwth-aachen.de}, Kopernikusstr. 16, 52074 Aachen, Germany}
		
		\title{CNN Cascades for Segmenting \\Whole Slide Images of the Kidney}
			
			\begin{abstract}
				\noindent
				%\textbf{Background:}
				Due to the increasing availability of whole slide scanners facilitating digitization of histopathological tissue, there is a strong demand for the development of computer based image analysis systems.
				In this work, the focus is on the segmentation of the glomeruli constituting a highly relevant structure in renal histopathology, which has not been investigated before in combination with CNNs.
				%
				%\noindent
				%\textbf{Methods:}
				%As CNNs were not yet applied to this segmentation task, we evaluate a state-of-the art method..
				We propose two different CNN cascades for segmentation applications with sparse objects. These approaches are applied to the problem of glomerulus segmentation and compared with conventional fully-convolutional networks.
				%To segment these regions-of-interest, we investigate different deep neural network architectures constituting the current state-of-the-art for supervised segmentation in biomedical applications.
				%Apart from single networks, we specifically propose and investigate two multi-stage approaches relying on two networks applied on different resolutions and basing on different training data.
				%
				%\noindent
				%\textbf{Results:}
				Overall, with the best performing cascade approach, single CNNs are outperformed and a pixel-level Dice similarity coefficient of 0.90 is obtained.
				Combined with qualitative and further object-level analyses the obtained results are assessed as excellent also compared to recent approaches.
				%
				%\noindent
				%\textbf{Conclusions:}
				In conclusion, we can state that especially one of the proposed cascade networks proved to be a highly powerful tool for segmenting the renal glomeruli providing best segmentation accuracies and also keeping the computing time at a low level. % Best combination?
			\end{abstract}

			
	\end{frontmatter}

\newcommand{\LINEWIDTH}{12cm}

\begin{figure*}[bth]
	\centering
	\includegraphics[width=\linewidth]{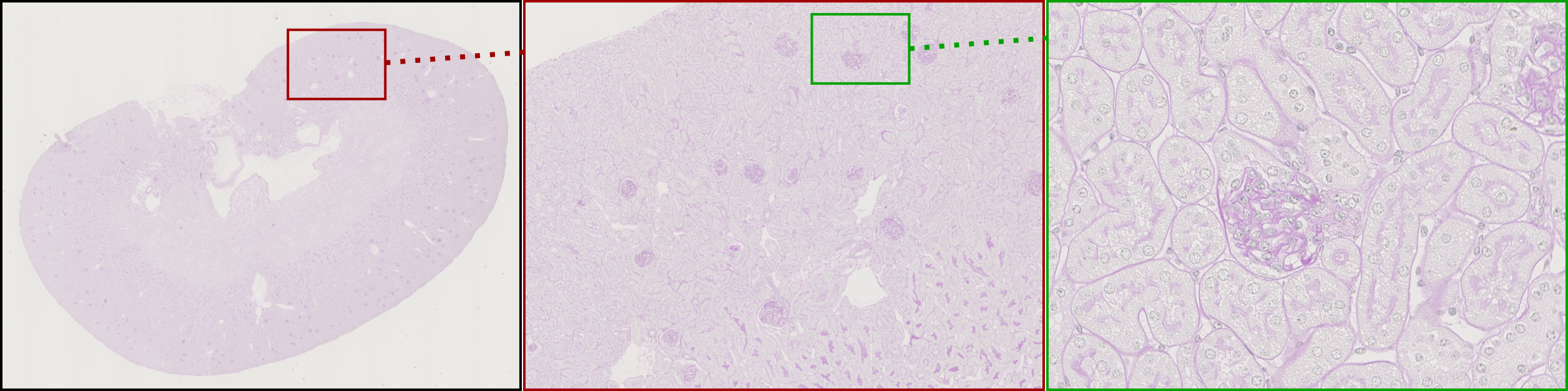}
	\caption{Example WSI showing a complete mouse kidney (left) as well as magnified versions of a selected patch. The right patch exhibits one of the regions-of-interest, a so-called glomerulus.}
	\label{fig:example}
\end{figure*}

%\linenumbers

%TODO are refs ok? cite, citep

\section{Introduction}

% problem definition
The digitization of histopathological tissue by means of so-called whole slide scanners results in highly resolved image data exhibiting resolutions in the range of gigapixels. As a consequence, the resulting whole slide images simultaneously contain large context (e.g. a slice of a mouse kidney, Fig.~\ref{fig:example}, left) and a high level of detail (Fig.~\ref{fig:example}, right).
%
%In histopathological research generally large image data need to be analysed. This is due to the highly resolved so-called whole slide images (WSIs) which exhibit resolutions in the range of one gigapixel or even more.
%
% what is the scope
In this work, we focus on an application scenario
from kidney pathology, specifically, we aim at segmenting the so-called glomeruli (Fig.~\ref{fig:example}, right) which are potentially the most relevant structures in histopathological research of the kidney (see Sect.~2). In this study, focus is on mouse kidney which is particularly relevant in research. However, mammalian kidneys in general are highly similar and therefore it can be assumed that an approach developed for mouse kidney can be also applied to other mammalian kidneys after small adjustments.

% manual issues
Due to the large size of renal WSIs, a manual annotation (detection and encircling) for research purposes is cumbersome and extremely time consuming (Fig.~\ref{fig:example2}). Apart from the difficulty of detecting each of the small objects (diameters of approx. 300 pixels) in the huge WSIs, the segmentation procedure is also subject to inter-observer variability showing potential for bias in the evaluations. A further issue is given by the variable sizes of the objects primarily because the 3D objects are cut at different positions and not necessarily centrally. During manual annotations scientists have to decide whether the respective glomerulus is large enough to be included into the statistics (as small objects are generally not considered) also providing potential for bias in the evaluation.

% automated segmentation
Although histopathological image analysis is currently of high interest~\citep{myBarker16a,myBenTaieb16a,myHou16a,myRonneberger15a,mySertel09a,myVeta16a}, there exists few literature on methods for automated segmentation of renal WSIs (Sect.~\ref{sec:related}).
One inhibiting factor for the development of segmentation methods is given by the relatively high initial effort required for annotating large amounts of training and testing data which are required for most state-of-the-art approaches.
This is especially time consuming because, due to the relatively small area covered by the sparsely occurring glomeruli, large overall data need to be manually processed~\citep{myGadermayr16f}.
Moreover, difficulties for automated segmentation approaches arise because of a high variability due to the staining process, due to variations in the cutting process as well as inter-subject variability~\citep{Gadermayr16e}.
Especially in case of pathological tissue, the outline of the glomeruli as well as the texture of the tissue can change dramatically.

Convolutional neural networks (CNNs) generally exhibited excellent performance in recent work on biomedical image analysis and especially in the field of histopathological image analysis~\citep{myBenTaieb16a,myRonneberger15a}. For segmentation tasks, especially fully-convolutional networks (FCNs) and derivatives~\citep{myBenTaieb16a,Girshick2014,Long2015,myRonneberger15a,Seyedhosseini2013} yielded excellent performances.
Previous work~\citep{Gadermayr17b} additionally identified them to be highly robust to staining variability providing further incentive for an investigation with respect to WSIs of the kidney.

However, the application addressed here distinctly differs from typical tasks for FCN-based segmentation~\citep{myBenTaieb16a,myRonneberger15a}. Whereas the original images are huge, the glomeruli, which should be segmented, are small objects also constituting a small overall area compared to the negative (non-glomerulus) class~(Fig.~\ref{fig:example}). Large areas in central kidney regions (medulla) even typically do not show any glomeruli.

\subsection{Medical Background}
%Mammalian kidneys have a complex anatomy and virtually all kidney diseases cause structural changes which currently constitute the most specific biomarkers. For a correct diagnosis and thereby also for an adequate therapy of patients, a histological evaluation of kidney biopsies is therefore essential. 
Histological analyses of kidneys are crucial and are broadly utilized in basic and translational research, in drug development as well as in drug toxicity testing. 
Morphological and ultrastructural alterations provide information on the pathomechanisms of renal injury, location (which compartment and/or which cells of the kidney are affected), distribution (is the damage focal or diffuse or segmental or global within a glomerulus) as well as severity of damage. Glomeruli are the blood-filtration units and are affected in a large number of kidney diseases. Analyses of glomeruli is therefore definitely one of the most important steps in histological evaluation of kidney biopsies in both clinical as well as experimental nephropathology. To obtain morphological information of specific structures of interest, different histological stainings are applied. The most important staining in renal histopathology is periodic acid schiff (PAS) providing important information for diagnosis (Fig.~\ref{fig:example}, Fig.~\ref{fig:example2}).
%Fig.~\ref{fig:example} shows an example WSI stained with PAS.

\subsection{Related Work} \label{sec:related}

%The problem of glomerulus segmentation has been recently addressed by various different approaches:
Few approaches were proposed to perform segmentation of the glomeruli.
\cite{myKato15a} introduced a separate detection and segmentation stage, both based on training a supervised classification model, specifically a linear support vector machine. For detection and segmentation different versions of histogram of oriented gradient descriptors were applied. Detection was performed by means of the sliding window approach. For segmentation, a polygon-fitting technique was utilized to determine the precise outline of the objects by adding a circularity constraint.
For learning to distinguish between border region and non-border regions (and not between inside and outside of a glomerulus), this method requires highly precise annotations for training. This method was developed for a specific staining (desmin) which highlights cells in glomeruli.
%Additionally, this method was developed for a specific staining (desmin) which highlights cells in glomeruli, facilitating a detection and segmentation. Experiments on the detection of glomeruli with the more common periodic acid schiff~(PAS) staining in combination with limited training data did not lead to suitable outcomes~\citep{Gadermayr16f}.

\begin{figure}[bt]
	\centering
	\includegraphics[width=\linewidth]{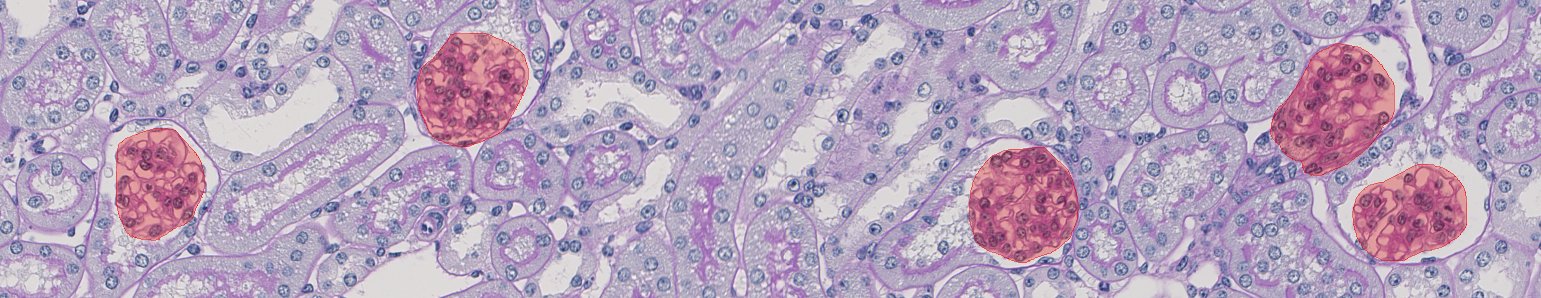}
	\caption{Example ground-truth annotations for several glomeruli in a renal WSI.}
	\label{fig:example2}
\end{figure}

To circumvent the need for large annotated training data, \cite{mySamsi12a} proposed an unsupervised segmentation algorithm. The authors aim at segmenting the Bowman's capsule surrounding the glomerulus by applying a color-based segmentation approach. The segmentation method consists of k-means clustering after thresholding including morphological operations. The inherent problem here is, that the Bowman's capsule is not always visible, depending on the staining protocol as well as the cutting plane~(Fig.~\ref{fig:example},~\ref{fig:example2}). Therefore, this method is not capable of generating a reliable precise segmentation of all objects.

All further approaches do not focus on a segmentation but on related tasks such as tissue classification or object detection:
In the work of~\cite{Gadermayr16f} the need for large annotated training data was circumvented by a weakly-supervised approach. It was shown that even with a small number of positive training samples, a detection of the glomeruli can be performed. However, a segmentation stage is missing in this work.
In a study of \cite{myHerve11a} various image representations were investigated to determine whether they are appropriate for distinguishing between different regions in renal pathology. The authors found that colour as well as texture representations can be utilized for distinguishing between the investigated texture classes.
In a further study~\citep{Gadermayr16e} domain adaptation methods were investigated for patch-wise classification of renal tissue. It was shown that domain adaptation can help to improve the fit of trained models for image data showing changed properties. However, the performance of feature-based patch-wise classification was not sufficient for a reliable classification or segmentation applying a difficult-to-segment staining.

\cite{myPedraza17a} applied a CNN for distinguishing between glomerulus and non-glomerulus texture based on previously extracted patches (similar to \cite{Gadermayr16e,myHerve11a}). Although this application is distinct from a segmentation, it provides evidence for the principal effectiveness of CNNs for distinguishing between glomerulus and non-glomerulus tissue.
Neural networks were also applied by~\cite{myLedbetter17a} with focus on a regression on WSI level. Specifically, the authors focussed on directly predicting the kidney function based on biopsies from patients suffering from chronic kidney disease without previously segmenting the glomeruli.

%other than renal
Besides renal pathology, CNNs were effectively applied in further histopathological image analysis applications. Classifications of WSIs with different targets were performed by~\cite{myBarker16a,myHou16a,mySertel09a}.
CNNs also yielded excellent performances in segmenting and classifying histopathological image data~\citep{myBenTaieb16a,myVeta16a}.
For segmentation, besides the sliding-window-based approach~\citep{NIPS2012_4741}, especially FCNs as well as modifications were successfully applied in the field of biomedical image analysis~\citep{myBenTaieb16a,myRonneberger15a} especially due to their high efficiency.

Since the work of \cite{myViola01a}, cascades became highly popular, specifically in the field of face detection.
Cascades were also combined with state-of-the-art CNN architectures and were not only applied to face detection~\citep{myLi15c} but also for segmentation applications~\citep{myDai16a,myJackson16a,myMa17a}.
\cite{myJackson16a} proposed a method to improve facial part segmentation performance by first detecting facial landmarks. 
\cite{myMa17a} introduced a cascade for combining segmentation and classification of thyroid nodules in ultrasound images.
\cite{myDai16a} proposed a CNN cascade including feature sharing to separately address detection, segmentation and classification.
In contrast to these multi-task approaches~\citep{myDai16a,myJackson16a,myMa17a}, here the focus is on segmentation only.
%The targets are diverse, such as multi-task learning~\citep{myDai16a} or a separation of segmentation and classification~\citep{myMa17a,myJackson16a}.

% task delegation effective?
% 

% first to address segmentation of kindey
% answer: are cascade CNNs more effective
% two different cascade CNN approaches proposed and evaluated

\subsection{Contribution}
We propose two CNN cascades for segmentation applications with sparse objects-of-interest.
By applying cascade approaches, one network is optimized for detection (i.e. a rough segmentation) based on the whole input images and another network is tweaked for a precise segmentation based on the output of a detection.
The introduced approaches are evaluated with respect to the segmentation of glomeruli in WSIs of mouse kidneys and are compared to conventional single CNNs trained with different strategies.
This application scenario was not investigated before in combination with CNNs.
Compared to typical segmentation scenarios for CNNs, the objects-of-interest in this task are small, sparse objects also constituting a small relative overall area.
Exhaustive evaluation is performed on pixel- as well as object-level figuring out positive and negative aspects of the different approaches.

The remainder of this paper is structured as follows: Section.~2 provides details of the methods proposed and investigated in this paper.
In Sect.~3, the experimental setting as well as the results are shown. In Sect.~4, the results are discussed and finally, Sect.~5 concludes this paper.

\section{Methods} \label{sec:main}

We investigated four different segmentation pipelines consisting of two conventional single FCNs as well as two cascade networks (Sect.~\ref{sec:main}.1). The individual network architectures consisting of the original and a shallow version of the U-Net~\citep{myRonneberger15a} (U-Net-D and U-Net-S) and a sliding window CNN method~\citep{NIPS2012_4741} (SW-CNN) are provided in Sect.~\ref{sec:main}.2. Further details on data augmentation, experimental details, image data and implementations are provided in Sect.~\ref{sec:main}.3-\ref{sec:main}.6.

%We implemented and trained three different network architectures (Sect.~\ref{sec:main}.1) and investigated several segmentation pipelines consisting of combination of the networks (Sect.~\ref{sec:main}.2). Further details on data augmentation, experimental details, image data and implementations are provided in Sect.~\ref{sec:main}.3-\ref{sec:main}.6.

\subsection{Segmentation Pipelines}

%We proposed and evaluated four different pipelines consisting of single networks as well as combinations of neural networks.
%Apart from two single networks, we propose two multi networks with separate detection and segmentation stages:
%The combination is motivated by the fact that only a small part of the WSI is covered by glomeruli (approx. $2 \% $). Furthermore, all glomeruli are located in the cortex (outer part of the kidney) whereas the medulla (inner part) usually does not contain any of them.
%
%By applying a multi-network approach, one network can be optimized for a detection (i.e. a rough segmentation) based on the whole input images and another network can be tweaked for a precise segmentation based on the output of a detection.

%Additionally, the sliding window approach applied for segmentation is highly inefficient, however, as a first detection stage it could be efficiently applied on a lower image resolution.

%To exploit (1)~contextual information for a rough segmentation (detection), (2)~to facilitate an accurate fine segmentation stage and (3)~to simultaneously increase efficiency, we investigate the following four configurations:
%TODO I think this is still not really consistent

The following four different segmentation pipelines based on different training strategies are investigated in this work:

\subsubsection{Single Net 1 (SN1)}
	For this approach, the U-Net-D is trained on the high resolution image data~(for explanations of the resolutions see Sect.~\ref{sec:imagedata}). To balance the ratio between positive and negative pixels, training is performed with patches randomly sampled with the restriction that at least a part of an object (glomerulus) must be visible. Due to the sparse occurrence of the objects, positive pixels still occur less often.
	The U-Net architecture was chosen due to the efficient end-to-end learning and the high performances obtained in biomedical applications~\citep{myRonneberger15a}.
\subsubsection{Single Net 2 (SN2)}
	For this approach, the same network architecture (U-Net-D) and image resolution is applied.
	Training is performed with data randomly sampled all over the kidney in order to uniformly incorporate data from regions distant to the objects-of-interest. This is also motivated by the fact that the negative class shows more variability compared to the positive (glomerulus) class~\citep{Gadermayr16f}. %However, the class labels are thereby highly imbalanced.
\subsubsection{Cascade Net 1 (CN1)}
	It can be assumed that training in an extremely class-imbalanced sense (SN2) is not beneficial for precise segmentations whereas an imbalance considering data variability (SN1) could lead to false detections.
	To circumvent a trade-off between precision and recall, we propose two cascade network approaches. CN1 consists of the SW-CNN on the low resolution image for a detection of the regions-of-interest followed by a precise segmentation of the U-Net-D on the high resolution image.
	The U-Net-D is only applied to patches showing detected objects and training is performed for both networks individually.
	To tweak the two networks for their purposes, the SW-CNN is trained based on randomly sampled data (as in case of SN2), whereas the U-Net-D is trained on object containing data only (as in case of SN1).
\subsubsection{Cascade Net 2 (CN2)}
	CN2 consists of the shallow U-Net-S applied to low resolution images followed by a deep U-Net-D applied to the high resolution data.
	The shallower U-Net-S compared is utilized for the first stage in order to account for the lower image resolution.
	It is trained based on randomly sampled data (as in case of SN2), whereas the U-Net-D is trained on object containing data only (SN1).
	As in case of CN1, the U-Net-D is only applied to regions showing detected objects.
	The application of an FCN in a first stage is motivated by the high computational efficiency. 
	Furthermore, the impact of the different architecture compared the SW-CNN should be assessed.

\subsection{Neural Network Architectures}
Three different network architectures were investigated consisting of a pixelwise CNN relying on sliding window segmentation and two different FCNs. 
All of them proved to be highly effective for biomedical segmentation applications.

\subsubsection{Sliding-Window Convolutional Neural Network (SW-CNN)}\label{sec_pixelwiseCNN}
The first network architecture is given by the segmentation approach introduced by \cite{NIPS2012_4741} which is based on a sliding window approach using a CNN for classification on patch-level.
The utilized network structure consists of alternations between convolutional (three $4\times4$ and one $5\times5$ layers) and pooling layers ($2\times2$) followed by one fully connected layer with 200 neurons and the output layer containing two neurons, one for each class (glomerulus and non-glomerulus) as proposed by~\cite{NIPS2012_4741}. The input image size is $95\times95$ pixels.

\begin{figure}[bt]
	\centering
%	\small
%	\setlength{\tabcolsep}{2.5pt} 
%	\begin{tabular}{cccc}
	%	\subfloat{\includegraphics[width=.24\textwidth]{CNNs/checkerboard_original.png} 
	%	\includegraphics[width=.24\textwidth]{CNNs/checkerboard_foveation.png} &
	%	\includegraphics[width=.24\textwidth]{CNNs/checkerboard_nonUniSampling.png} &
		%\includegraphics[width=.24\textwidth]{CNNs/checkerboard_nonUniSamplingAndFoveation.png}
		\subfloat[Original]{\includegraphics[width=.24\textwidth]{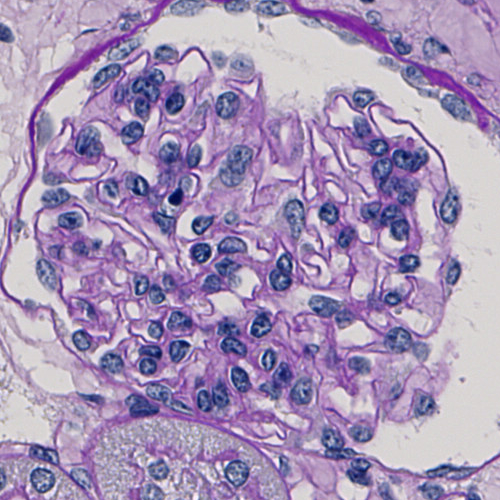}} \hspace{0.01cm} %&   
		\subfloat[Foveation]{\includegraphics[width=.24\textwidth]{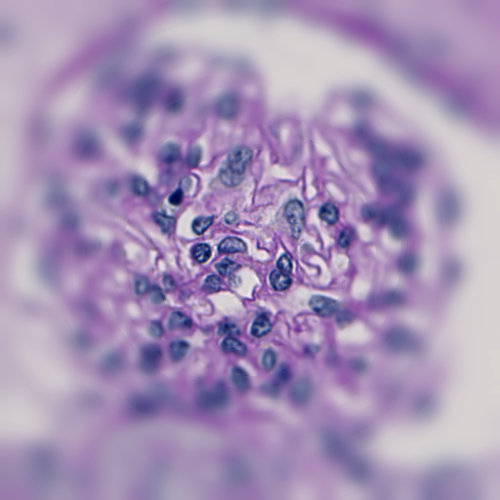}}\hspace{0.01cm} %& 
		\subfloat[Non-uniform sampling]{\includegraphics[width=.24\textwidth]{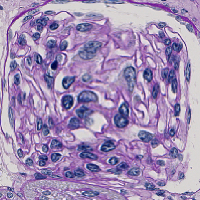}} \hspace{0.01cm}
		\subfloat[Foveation and non-uniform sampling]{\includegraphics[width=.24\textwidth]{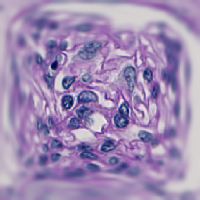}}
	%	\textbf{(a)} original & \textbf{(b)} foveation &  %\multirowcell{2}{\textbf{(c)} non-uniform\\sampling} & \textbf{(d)} both\\
%		& & &
%	\end{tabular}
	\caption{Effect of foveation and non-uniform sampling on a glomerulus patch.}
	\label{foveationAndNonUniformSampling}
\end{figure}

We also applied foveation \citep{NIPS2012_4741} utilizing different Gaussian kernels ($\sigma \in [1, 9]$) and non-uniform sampling to provide the network with more context information while keeping the input image size small (Fig.~\ref{foveationAndNonUniformSampling}).
%Displacement vectors $d$ (distance from the center) are computed such as $d = \frac{3x^3 + x}{4} \frac{n-n'}{2}$, where $n$ is the original patch size and $n'$ is the new patch size and $x$ is the offset ($x \in [-1, 1]$) from the center in the original patch (applied similar for horizontal and vertical coordinates). Finally $d$ is rounded and added to the coordinates.

\subsubsection{Fully Convolutional Networks (FCN)}\label{sec_UNet}
\cite{Long2015} introduced an FCN that is able to generate output pixel maps instead of single pixels only. This so-called end-to-end training especially increases efficiency of segmentation compared to the sliding window approach~\citep{NIPS2012_4741}.
We investigated a modified version of the traditional FCN, the so-called U-Net~\citep{myRonneberger15a} which combines up-sampled features from deep, coarse layers with those from shallow, fine layers and exhibited excellent performances in biomedical applications. Similar architectures were also proposed in literature~\citep{Girshick2014,Long2015,Seyedhosseini2013}.

%The convolutions do not use any padding leading to an output mask which is smaller by $184$ pixels in width and height. After each max pooling step the number of filters is doubled up to 1024 filters in the lowest resolution. 
The contracting path of the U-Net consists of two $3\times3$ convolutions followed by a $2\times2$ max pooling. The expanding path consists of an up-sampling step, followed by a $2\times2$ convolution, during which the number of feature channels is divided by two. The feature maps are then combined with the feature maps of matching resolution from the contracting path facilitating training with view training samples and/or allow to use deeper network architectures. 

Besides the original network architecture~\citep{myRonneberger15a} consisting of 23 convolutional layers (U-Net-D), we furthermore investigated a shallower architecture consisting of 13 convolutions to take account of different image resolutions (U-Net-S).
A comparison of the networks investigated in this work is provided in Table~\ref{tab_techDiffNets}.

\begin{table}[htb]
	\centering \footnotesize
	\renewcommand{\arraystretch}{1}
	\begin{tabular}{l|ccrccc}
		network 		& \makecell{input\\size}& \makecell{output\\size} & \makecell[c]{network\\parameters}  &\makecell{conv.\\layers}& \makecell{pooling\\layers} & \makecell{training\\time}\\ \hline   
		SW-CNN  &  $95\times95$  	&   $1\times2$  	&220,826		& 4  & 4 & 0.5\,h\\
		%	FCN  	& $495\times495$  	&  $28\times28$   &133,873  	& 5  & 4 &0.7\,h\\
		U-Net-D	& $492\times492$  	&  $308\times308$   &31,031,745   & 23 & 4 &5.5\,h \\
		U-Net-S	& $492\times492$    &  $452\times452$   &1,862,849    & 13 & 2 &4.5\,h
	\end{tabular}
	\caption{Comparison of the investigated different network architectures.}
	\label{tab_techDiffNets}
\end{table}

\subsection{Data Augmentation \& Training Data}
As the glomeruli do not have a distinct orientation, rotation (multiples of $90^\circ$) and flipping was performed without losing any distinctive information.
Additionally, we applied elastic deformations~\citep{myRonneberger15a,bestPractice,whenToWarp} to cover natural variability.
In the case of the SW-CNN, additionally foveation and non-uniform sampling was applied as described in Sect.~\ref{sec_pixelwiseCNN}.

\subsection{Experimental Details}
As common practise for preprocessing, we transformed the data to show zero-mean and unit-variance.
For training the networks, batch normalization was applied in combination with L2 Regularization.
The batch size was set to five for the FCNs and to 100 for the SW-CNN.
All networks were trained using cross-entropy as a loss function and Adam~\citep{Kingma2014} as an optimizer with learning rate $\nu = 10^{-5}$.
The networks were trained for 15 epochs, where the optimum epoch was evaluated based on the validation error. The reported rates were obtained on a separate test set.

In order to facilitate unbiased evaluation and to obtain an outcome for each WSI, eight-fold cross validation was applied. The available 24 WSIs were split into training, validation and test set containing 18, three and three images, respectively. 
For the U-Net architectures, we extracted 2,700 patches from the 18 training WSIs for each fold for training. These patches where further augmented resulting in 10,800 patches for the U-Net-D and 27,000 patches (due to the smaller input image size) for the U-Net-S.
For the SW-CNN, overall 864,000 augmented samples were extracted from 3,600 original patches.
For the sliding window approach (SW-CNN), a step size of five pixels (in the downscaled image) was selected to keep the computation effort comparable to the FCN.

\subsection{Image Data} \label{sec:imagedata}

The WSI data was obtained from resected mouse kidneys.
Kidneys were processed as previously described by~\cite{myBoor15a}. They were fixed in methyl Carnoy's solution and embedded in paraffin. Paraffin sections (1 $\mu m$) were stained with PAS reagent and counterstained with hematoxylin. Whole slides were digitalized with a Hamamatsu NanoZoomer 2.0HT digital slide scanner in combination with a 20$\times$ objective lens.

The investigated overall image data set consisted of 24 WSIs showing mouse kidney stained with PAS with sizes of roughly $50 000 \times 40 000$ pixels.
All of these slides were manually annotated, specifically all glomeruli were encircled by a polygon by a student helper\footnote{Special thanks to Hanry Ham for supporting us with highly precise annotations} supervised by an expert (coauthor B.~M.~Klinkhammer).
As in previous work~\citep{myKato15a}, the second highest resolution (i.e. the highest resolution downscaled by factor 2) was accessed for precise segmentation, constituting the "high resolution". Additionally, for first-stage segmentation (CN1, CN2) we also made use of a resolution obtained by additionally downscaling by factor 4 (referred to as "low resolution").
Due to the multi-resolution image format (Hamamatsu ndpi), these resolutions can be efficiently directly accessed without the need for actual downsampling.

\subsection{Implementation Detail}
All approaches were coded in python using Keras~\citep{keras} as a high-level neural networks API, running on top of TensorFlow~\citep{tensorflow}.
Evaluations were performed on NVIDIA TITAN X graphics cards.

\section{Results}

Evaluation was first performed on pixel-level (Sect.~\ref{sec:resPix}), i.e. the binarized CNN output was interpreted as class label map. These output label maps were directly compared with the ground-truth label maps for the complete WSIs.
Furthermore, the segmentation performances were evaluated on object-level (Sect.~\ref{sec:resObj}). For this purpose, we computed the performances individually for each glomerulus and finally aggregated the outcomes by computing the cumulative distributions based on the segmentation metric.

Whereas pixel-level analysis facilitates assessing the general performance considering false classifications, object-level analysis additionally allows to separate a false detection of objects and an imprecise segmentation.

\subsection{Pixel-Level Analysis} \label{sec:resPix}

\begin{figure} \center \includegraphics[width=7cm]{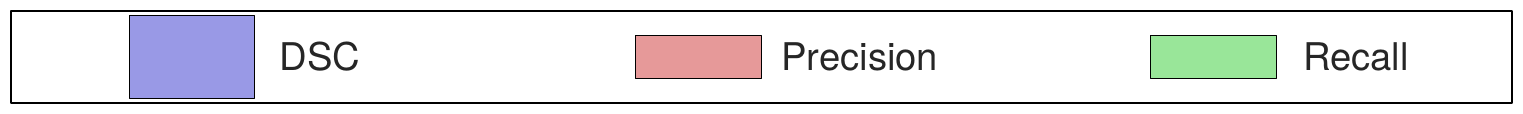}
	\subfloat[Original CNN output without postprocessing]{\includegraphics[clip=true, trim=1.5cm 8cm 1.5cm 8cm, height=4.85cm]{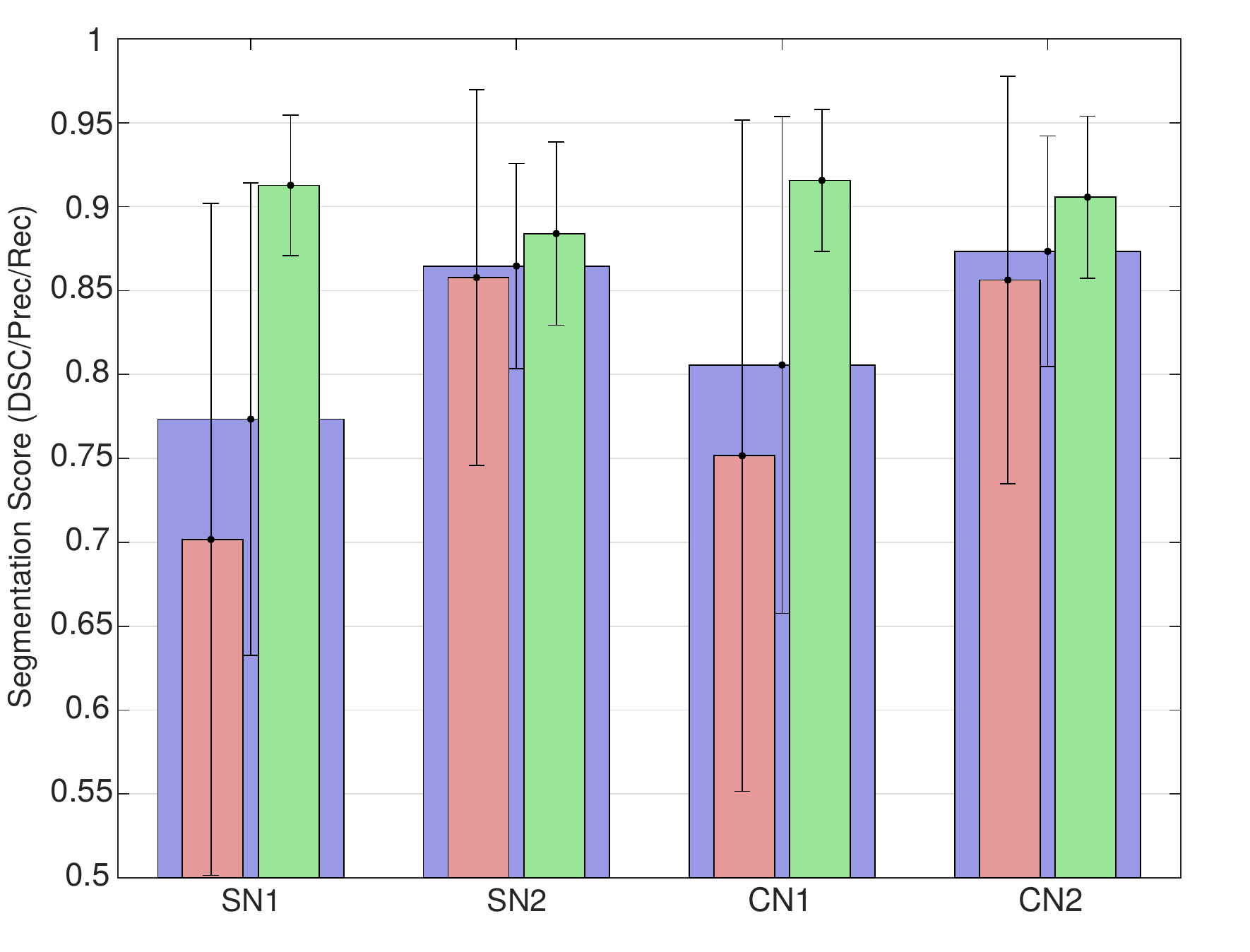}\label{fig:res1a}} 
	\subfloat[Domain specific evaluation]{\includegraphics[clip=true, trim=3.15cm 8cm 1.5cm 8cm, height=4.85cm]{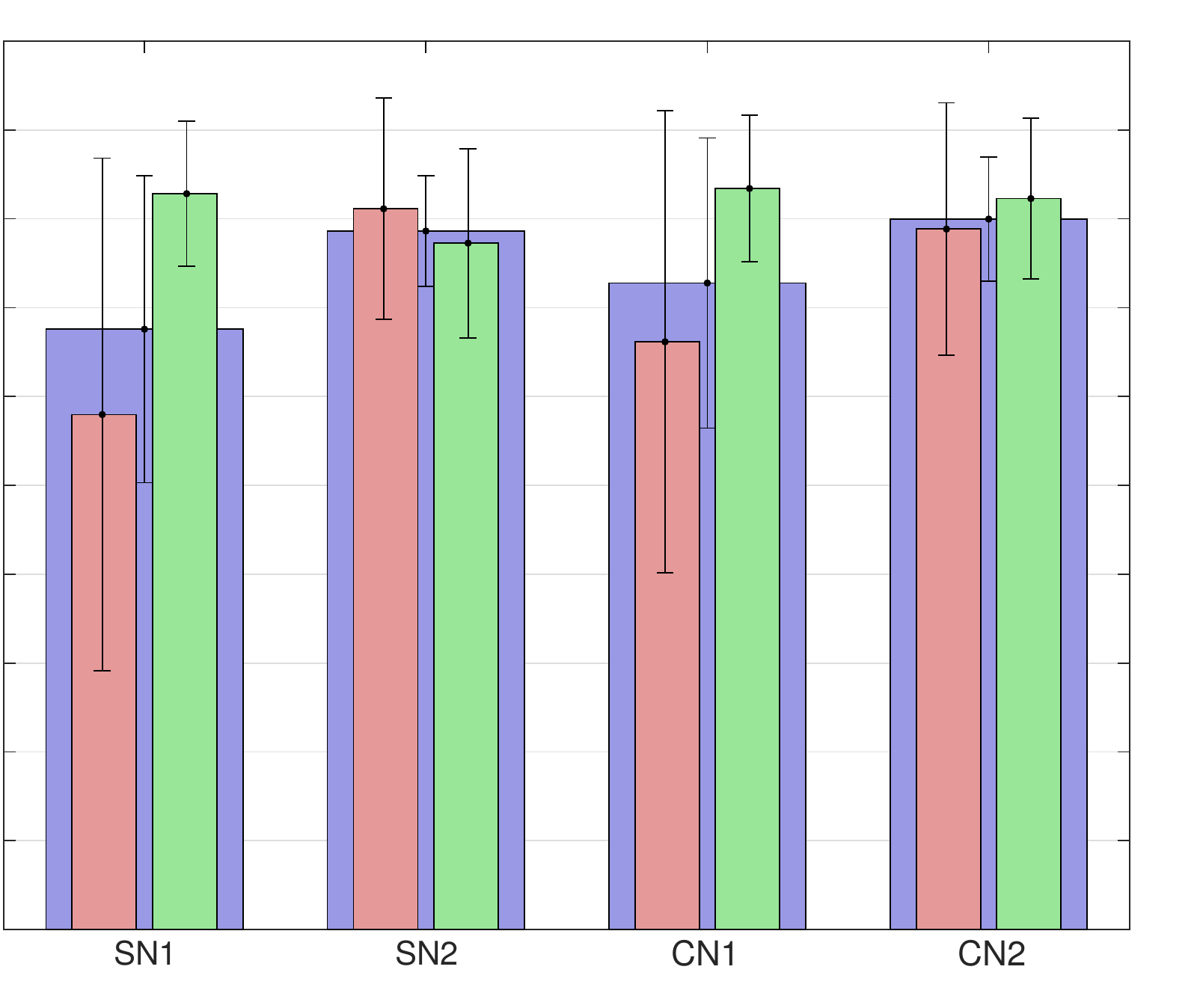}\label{fig:res1b}}
	\caption{Results (DSC, precision and recall) of pixelwise classification for the four investigated approaches.}
	\label{fig:res1}
\end{figure}

For evaluating the segmentation performances, first the pixelwise classification performance was assessed. In Fig.~\ref{fig:res1a}, Dice similarity coefficient (DSC), precision and recall is provided for all individual approaches. Specifically, we provide mean and standard deviation over the individual WSIs. 

Considering the outcomes, we notice that balanced-training-based SN1 delivered DSCs of 0.77 and was distinctly outperformed by SN2 (DSC: 0.86) relying on data sampled all over the kidney although recall was decreasing.
With both multi-stage networks, CN1 and CN2, the precision of SN1 as well as the recall of SN2 were outperformed. Overall, DSCs were 0.81 and 0.87, respectively.

For further analyses (Fig.~\ref{fig:res1b}), we excluded glomeruli which were marginally cut only (and therefore smaller than 20,000 pixels) in the ground truth as well as in the segmentation. This is motivated by the fact that in research these objects are ignored as well. Furthermore, large objects (comprising more than 200,000 pixels) were deleted which were obviously no glomeruli (and can be easily removed automatically) but influenced the performance measure distinctly. This procedure can be understood as postprocessing.

Considering the thereby achieved measures, all of them increased and we obtained DSCs between 0.84 (SN1) and 0.90 (CN2). Especially precision increased distinctly for all approaches due to a reduced number of false-positive objects.
CN2 still showed the best performance followed by SN2 as in case of evaluating the original CNN outputs (Fig.~\ref{fig:res1a}).
%TODO which sizes

\subsection{Object-Level Analysis} \label{sec:resObj}

\begin{figure} \center
	\subfloat[Aggregated over all ground-truth glomeruli. False-positive objects are thereby neglected in this consideration.]{\includegraphics[height=4.7cm]{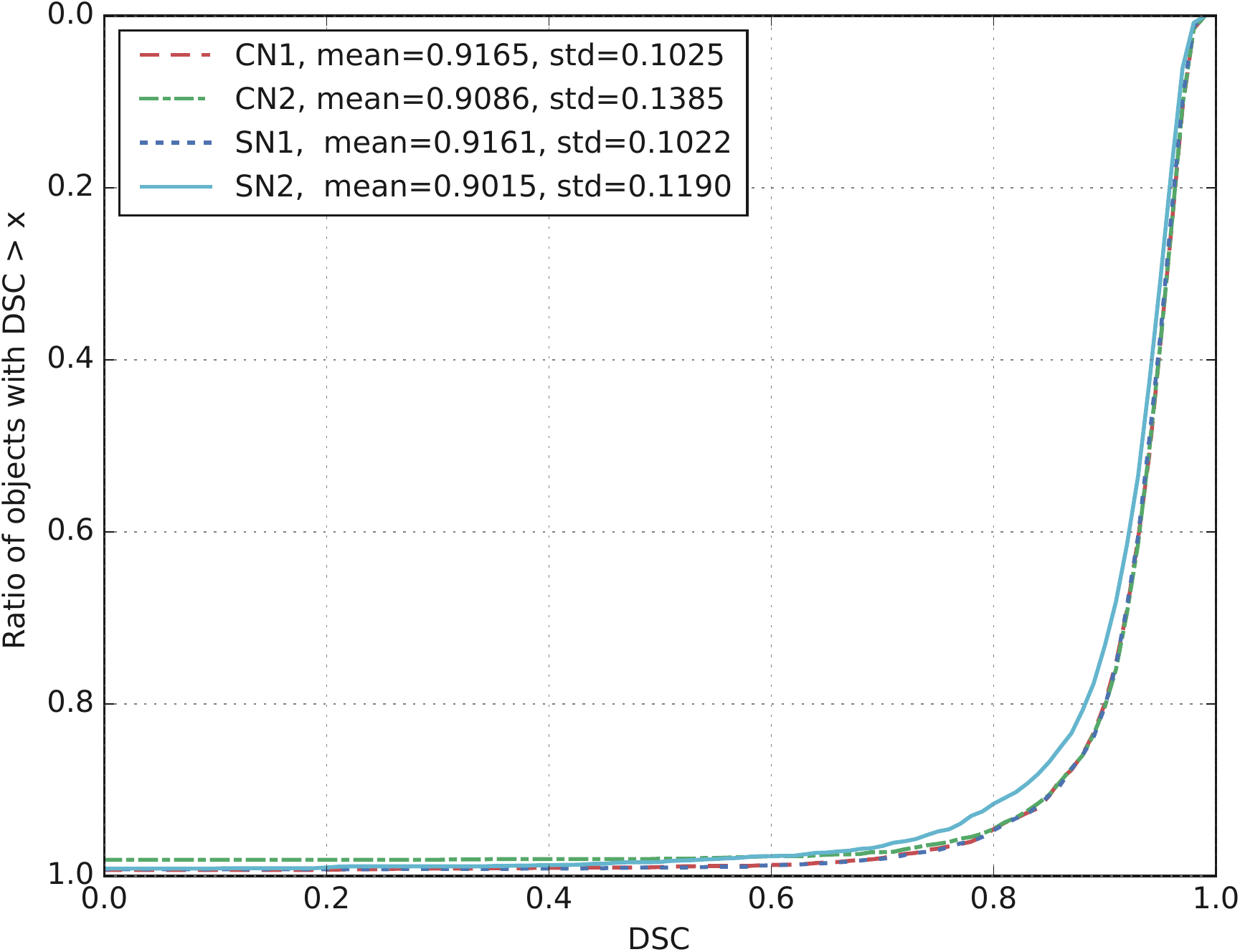}\label{fig:res2}} $\;$
	\subfloat[Aggregated over all detected glomeruli. False-negative objects are thereby neglected in this consideration.]{\includegraphics[height=4.7cm]{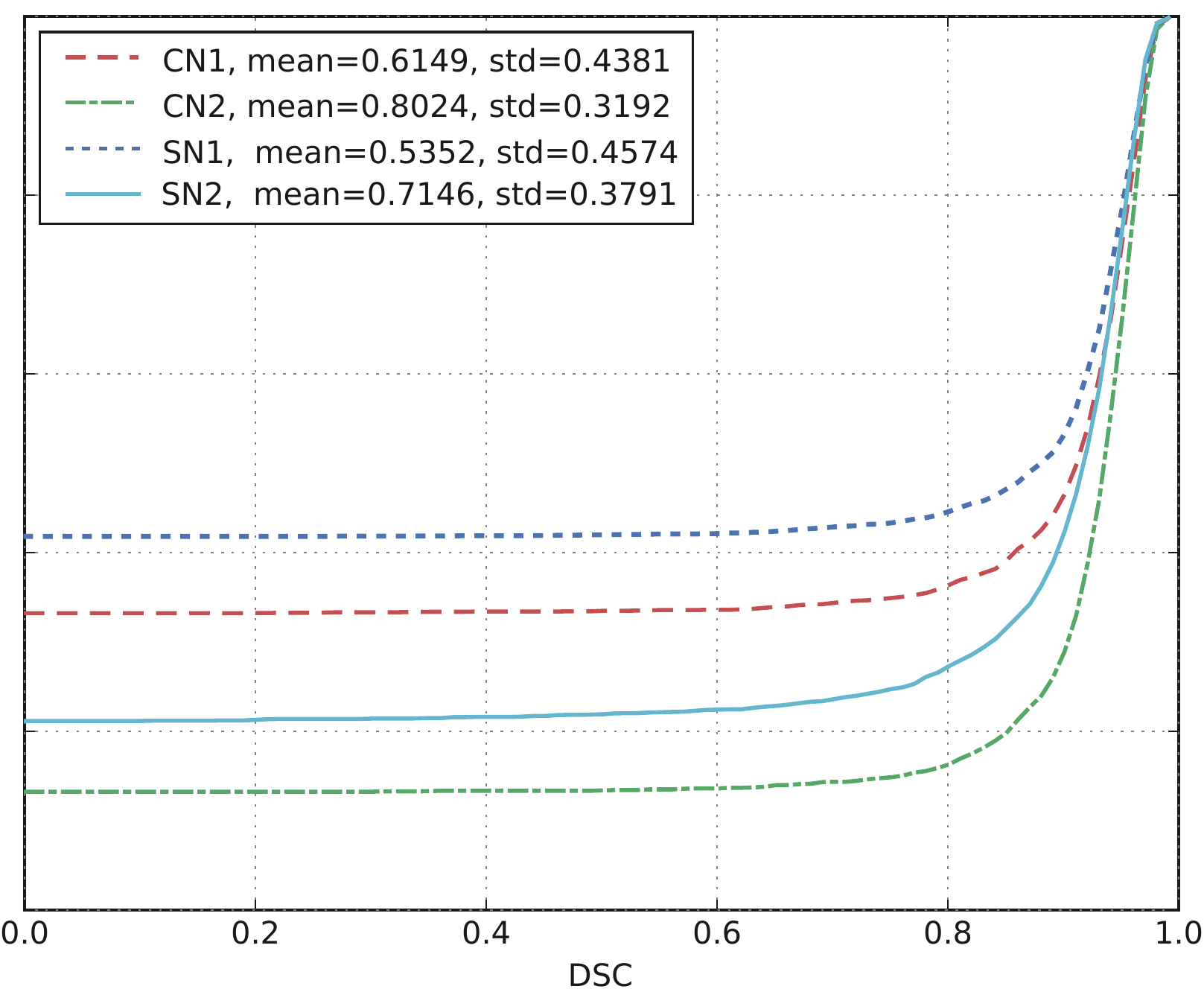}\label{fig:res3}}
	\caption{Object-level DSCs. These figures provide the cumulative distribution of DSCs aggregating all objects in the ground-truth (a) and aggregating all objects in segmentation output (b).}
\end{figure}

Furthermore, the segmentation performance was calculated on object level, i.e. we computed the performances (DSCs) individually for each glomerulus in the ground-truth and finally aggregated the outcomes by computing the cumulative distribution based on the measure as shown in Fig.~\ref{fig:res2}.

In this setting, CN1, CN2 and SN1 exhibited similar performances. 95 \% of ground-truth glomeruli showed a segmentation performance (DSC) of 0.80 and above.
80 \% of ground-truth glomeruli even showed a segmentation performance of 0.90 and above. The single resolution protocol SN2 showed worse classification rates.
%shown individually for the glomeruli 

This object-level analysis (Fig.~\ref{fig:res2}), however, completely ignores false positive objects as only the segmentation performance of ground-truth objects was aggregated and false positive objects were not assessed.
Therefore, we additionally analysed the performance based on the segmented objects ignoring that some objects were missed (i.e. false negative objects are neglected here). 

Fig.~\ref{fig:res3} shows the cumulative distributions of detected objects in a similar manner.
Whereas object-level recall was generally extremely high as almost each ground-truth object showed a DSC greater than 0.0 (left column in Fig.~\ref{fig:res2}), object-level precision differed more distinctly. 
SN1 and also CN1 exhibited the lowest precision (0.58 and 0.67) indicated by cumulative rates starting at a low DSC-level.
Especially CN2 showed the best rates over the whole DSC range. The obtained object-level precision was 0.86 (Fig.~\ref{fig:res3}, left column).

%\begin{figure} \center
%	\subfloat[]{\includegraphics[width=11cm]F1_all_gt_inkscape.pdf}}
%	%TODO use the correct figure
%	\caption{Object-specific DSCs aggregated over all detected glomeruli}
%	\label{fig:res3}
%\end{figure}

\subsection{Qualitative Analysis} \label{sec:resObj}

\begin{figure} \center
	\includegraphics[width=\linewidth]{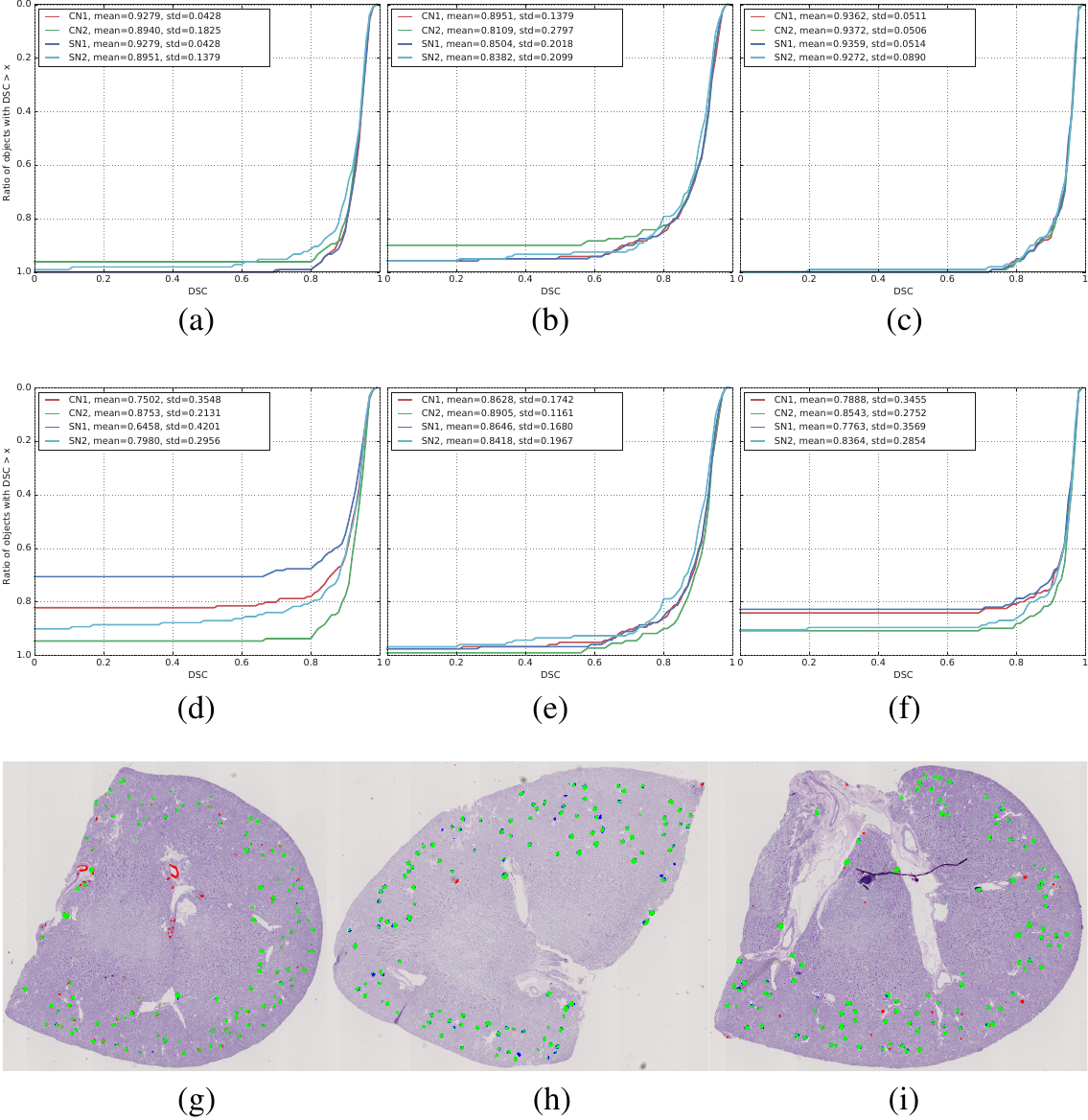}
	\caption{Example segmentation output (g)-(i) of three complete WSIs showing true-positives (green), false-positives (red) and false-negatives (blue) on pixel-level and the corresponding cumulative object-level results (correspondences: (a)-(d)-(g), (b)-(e)-(h), (c)-(f)-(i)).
	The top row (a)-(c) shows aggregations over all ground-truth objects (ignoring false-positives as in Fig.~\ref{fig:res2}) whereas the second row (d)-(f) provides aggregations over all detected objects (ignoring false negatives as in Fig.~\ref{fig:res3}). 
		}
	\label{fig:exoutput}
	%TODO improve figures (a) - (f)
\end{figure}

\begin{figure} \center
		\subfloat[Example segmentation output showing accurate segmentations]{\includegraphics[height=3.17cm]{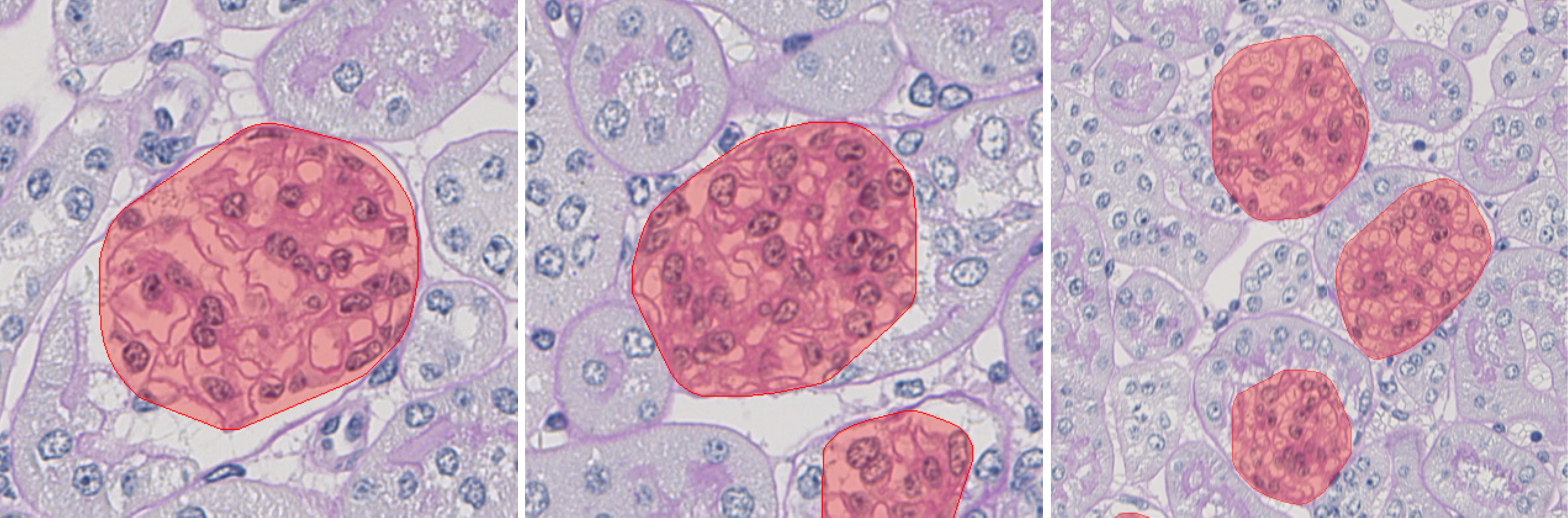}}\\
		\subfloat[Examples for mis-segmentations (too small segmentation, false negative object and false positive object)]{\includegraphics[height=3.17cm]{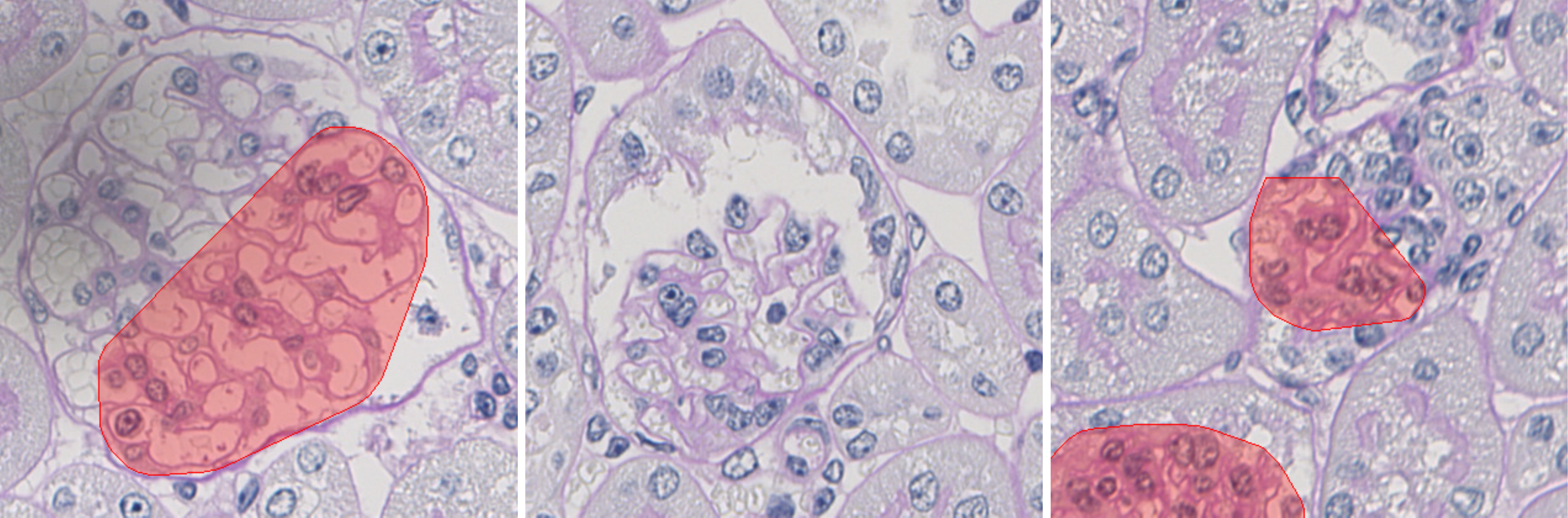}} $\;$
	\caption{Example segmentation output in a high resolution (red) extracted from the WSI shown in Fig.~\ref{fig:exoutput}h.}
	\label{fig:exoutput2}
\end{figure}

Exemplary segmentation outputs for complete WSIs are provided in Fig.~\ref{fig:exoutput} which also exhibits the cumulative object-level measures as summarized for all objects in Fig.~\ref{fig:res2} and Fig.~\ref{fig:res3}. %Whereas the WSI shown in Fig.~\ref{fig:exoutput}~(d) shows intermediate performance, (e) and (f) show a above-average number and (e) and a below-average number (f) of false positives.
Further example segmentation output on a high resolution is provided in Fig.~\ref{fig:exoutput2} showing examples for mis-segmentations as well as accurate segmentations.

\subsection{Runtime Evaluation}
Empty regions in the WSIs were previously detected by means of low-pass filtering, thresholding and morphological operations and thereafter ignored to increase efficiency.
The runtimes for the four pipelines are as follows: Whereas the single networks SN1 and SN2 showed similar execution runtimes of 135 s per WSI on average, CN1 and CN2 both exhibited improved efficiency resulting in 105 s (CN1) and 60 s (CN2) per WSI. The improvements of efficiency occurred due to fast first stages (CN1: 18 s, CN2: 3 s) distinctly decreasing the amount of segmented data in the second stage.

\section{Discussion}

%In this work, we investigated four different CNN architectures with respect to the segmentation of glomeruli in WSIs of mouse kidneys.
In this work, two CNN cascades were proposed, applied to the problem of glomerulus segmentation and compared with two conventional CNN approaches.

Considering the single network approaches SN1 and SN2, it can be observed that SN1 showed distinctly lower precision whereas SN2 exhibited slightly lower recall (Fig.~\ref{fig:exoutput}). This is obviously due to the divergent training setting with more negative data in case of SN2 compared to the rather balanced setting in case of SN1. In order to optimize the DSCs it can be summarized that the negative class exhibiting higher variability requires more training data in a single network setting.

Considering the cascade approaches it can be observed that both methods were able to increase the precision of SN1 while also increasing the recall of SN2. Training the first glomerulus detecting stage with large negative data while training the final segmenting stage with rather balanced data is obviously an effective approach to tackle a segmentation application with highly unbalanced class distributions.
Especially CN2, relying on two FCNs, is able to improve the overall segmentation performance (DSC) compared to a single FCN even further.
CN1, which is based on the sliding-window approach for detection of the glomeruli exhibited a lower precision also leading to a lower DSC. This is supposed to be due to the rather large step size which was chosen to keep the computational costs on a similar level compared to the fully convolutional network.
Although measures generally increase, the rankings are consistent if considering the original CNN output (Fig.~\ref{fig:res1a}) or the domain specific evaluation in Fig.~\ref{fig:res1b}. 

Considering the object-level scores (Fig.~\ref{fig:res2} and \ref{fig:res3}) it can also be noted that CN2 corresponds to the best overall approach providing the best outcomes for both evaluation procedures which is a highly interesting outcome as in general improved recall often corresponds to decreased precision and vice versa.
Comparing the object-level scores, we notice that CN2 was only negligibly inferior to any of the other approaches in the range between a DSC of 0.00 and 0.80 based on evaluating the ground-truth objects (Fig.~\ref{fig:res2}).
That means, some of the glomeruli which were weakly segmented (e.g. DSC 0.6 and below) with e.g. SN1 where not detected at all with CN2.
As the difference is minor and as it is questionable if weakly segmented objects are for the final problem not even more problematic than objects which were not detected at all, we conclude that these analyses identified CN2 as best approach.
Compared to pixel-level analysis, the object-level benefit of CN2 compared to SN2 is even more pronounced. This is supposed to be because in case of CN2 the second stage network was optimized for precise segmentation by training on more specific glomerulus-containing data only. 

To summarize, CN2 performed best on pixel-level as well as on object-level providing a highly appropriate ratio between precision and recall in both scenarios.
This approach also performed best considering the evaluation runtimes.
%This method generates the highest number of accurately segmented glomeruli (Fig.~\ref{fig:res2}) while also providing the fewest false positive objects (Fig.~\ref{fig:res3}).

\subsection{Comparison with Previous Work}
Due to little literature on glomerulus segmentation in combination with different investigated histological stainings, it is very difficult to compare the proposed approaches with previous work.

We can compare the detection performance with a previous sliding window detection method~\citep{Gadermayr16f} showing detection F$_1$-Scores of 0.80 for PAS stained WSIs. For CN2, we obtain detection rates of 0.91 (precision: 0.97, recall: 0.86) which is clearly higher. However, a direct comparison is not completely fair because in the recent work focus was on obtaining a small-data-based solution with very few training samples only.

An evaluation for glomerulus segmentation was performed by~\cite{myKato15a}. However, the authors focused on a immunostaining (desmin), which is a known glomerular injury marker specifically delineating cells in glomeruli with the disadvantage that not all glomeruli are explicitly highlighted~\citep{myKato15a}. As we focused on the general purpose PAS staining, a direct comparison of rates is not  equitable and also an application of the proposed method on our data is difficult.
Ignoring the different data and directly comparing the outcomes, it can be assumed that our method performs a more accurate segmentation as especially the object-detection rates of the proposed algorithm were lower (mean F$_1$-Score of 0.87) due to a larger number of false-positive. 
Considering the segmentation stage, 90 \% of the detected glomeruli showed an F-measure greater than 0.80 which was also higher in case of CN2.
%Although the final segmentation of true-positive objects is accurate (mean DSC of about 0.90), a strong limitation is given due to a low detection precision leading to low overall rates.

\subsection{Interpretation of Segmentation Performance}
The original evaluation on pixel level (Fig.~\ref{fig:res1}~(a)) was limited by the fact that the ground-truth annotation contains small objects which where not always annotated because at some point it becomes extremely difficult to determine if a marginally cut glomerulus actually is a glomerulus. As these small objects are not relevant for medical purposes, we excluded them (as well as very large objects with can be easily detected) for all further evaluations. The obtained outcomes are highly satisfying from a quantitative as well as qualitative point of view. We assume that the considered application scenario is consequently solved and focus in future should be on further issues such as variability due to pathologies and on different histopathological stainings.
Segmentation DSCs of 0.90 and above on pixel- and object-level indicate a good correspondence of ground-truth and segmentation output. As the ground-truth annotations are also subject to uncertainty and inter-observer variability, we assume that the obtained results are quite close to the limits of feasibility.

The obtained computing times of 60 s on average on a single GPU indicate a high effectiveness of the proposed pipeline which processes images with resolutions in the range of one gigapixel. Thereby, biomedical analyses can be performed without long delays.

\section{Conclusion}
Two CNN cascades were proposed, applied to the problem of glomerulus segmentation and compared with two conventional CNN approaches.
We can state that FCNs are generally highly powerful tools for this segmentation task. 
Specifically, a cascade consisting of an FCN for detection and another network for precise segmentation (CN2) delivered the best outcomes exhibiting a mean DSCs of 0.90 on pixel-level. Considering the results on object-level, 80 \% of the ground-truth objects were segmented with a DSC of 0.90 and above in combination with the fewest false-positives which is even more impressive and shows that this method can directly be utilized in practise for tissue showing no pathological variations.
We are confident that the proposed method can also applied to other problem domains with sparsely distributed objects-of-interest.

In future work, we will investigate the effect of typical pathologies which has not been performed so far. We will focus on iterative unsupervised domain adaptation of a healthy model by means of WSIs showing increasing severity. 

\subsection*{Acknowledgement}
This work was supported by the German Research Foundation (DFG) under
grant no. ME3737/3-1 to DM, and SFB-TRR 57 and BO 3755/3-1 both to PB.

\subsection*{Declaration of interest}
Conflicts of interest: none

\section{References}
\bibliography{literature,bib,/home/staff/gadermayr/bibtex/eigene,/home/staff/gadermayr/bibtex/my}

\end{document}